\documentclass{article}
\usepackage{ijcai25}

\usepackage{times}
\usepackage{soul}
\usepackage{url}
\usepackage[hidelinks]{hyperref}
\usepackage[utf8]{inputenc}
\usepackage[small]{caption}
\usepackage{graphicx}
\usepackage{amsmath}
\usepackage{amsthm}
\usepackage{amssymb}
\usepackage{algorithm}
\usepackage{algorithmic}
\usepackage[switch]{lineno}
\usepackage{subfigure}
\usepackage{graphicx}       
\graphicspath{{media/}}     
\usepackage{wrapfig}
\usepackage{enumitem}
\usepackage{multirow}
\usepackage{xcolor}
\usepackage{listings}
\usepackage{booktabs}
\usepackage{bera}

\colorlet{punct}{red!60!black}
\definecolor{background}{HTML}{EEEEEE}
\definecolor{delim}{RGB}{20,105,176}
\colorlet{numb}{magenta!60!black}
\usepackage{amsmath}

\DeclareMathOperator*{\argmin}{argmin}
\DeclareMathOperator*{\argmax}{argmax}

\title{ToolFactory: Automating Tool Generation by Leveraging LLM to Understand REST API Documentations}

\author{
    Xinyi Ni$^1$\and
    Qiuyang Wang$^1$\and
    Yukun Zhang$^1$\and
    Pengyu Hong$^1$\\
    \affiliations
    $^1$Computer Science, Michtom School of Computer Science , Brandeis University\\
    \emails
    \{xinyini, qiuyangwang, yukunzhang, hongpeng\}@brandeis.edu
}

\begin{document}

\maketitle

\begin{abstract}
LLM-based tool agents offer natural language interfaces, enabling users to seamlessly interact with computing services. While REST APIs are valuable resources for building such agents, they must first be transformed into AI-compatible tools. Automatically generating AI-compatible tools from REST API documents can greatly streamline tool agent development and minimize user learning curves. However, API documentation often suffers from a lack of standardization, inconsistent schemas, and incomplete information. To address these issues, we developed \textbf{ToolFactory}, an open-source pipeline for automating tool generation from unstructured API documents. To enhance the reliability of the developed tools, we implemented an evaluation method to diagnose errors. Furthermore, we built a knowledge base of verified tools, which we leveraged to infer missing information from poorly documented APIs. We developed the API Extraction Benchmark, comprising 167 API documents and 744 endpoints in various formats, and designed a JSON schema to annotate them. This annotated dataset was utilized to train and validate ToolFactory. The experimental results highlight the effectiveness of ToolFactory. We also demonstrated ToolFactory by creating a domain-specific AI agent for glycomaterials research. ToolFactory exhibits significant potential for facilitating the seamless integration of scientific REST APIs into AI workflows.

\end{abstract}

\section{Introduction}
Agents based on large language models (LLM)~\cite{Wang_2024} are rapidly proving their versatility and universal applicability in various fields, such as scientific research~\cite{baek2024researchagentiterativeresearchidea}, healthcare~\cite{abbasian2024conversationalhealthagentspersonalized}, finance~\cite{li2023tradinggptmultiagentlayeredmemory}, and so on. These agents show impressive reasoning and problem-solving capabilities, addressing real-world challenges. Tool agents, built on LLMs, aim at interacting autonomously with existing software or web services. For example, ChemCrow~\cite{bran2023chemcrowaugmentinglargelanguagemodels}, a tool agent recently developed for chemistry research, leverages domain-specific software to complete various tasks, such as searching, instruction, analysis, and prediction. However, constructing such agents often requires significant software engineering effort and domain knowledge. On the other hand, a tool agent can utilize existing \textbf{REST APIs}, especially those associated well-designed schemas that clearly explain in machine-understandable ways how the APIs should be used correctly and precisely. Developing such tool agents involves minimal programming effort. Mastering REST APIs enables tool agents to access a variety of data and computing services. For example, ToolLlama~\cite{qin2023toolllmfacilitatinglargelanguage} utilizes over 16,000 APIs from \href{https://rapidapi.com/hub}{RapidAPI}, creating a comprehensive toolkit to address various problems in multiple application domains. However, it relies on API platforms for generating API tools, which cannot be transferred to build research AI agent. 

Several obstacles hinder the development of tool agents in research domains. Unlike commercial APIs, scientific APIs often lack comprehensive documentation and rarely adhere to standardized schemas. Even when a schema is provided, it may be incomplete or missing critical information, complicating the creation of a generalized tool generation process. Additionally, scientific APIs and their documentation are typically updated less frequently than commercial counterparts, necessitating extensive validation and refinement before they can be effectively utilized.

We consider it essential to develop an automated pipeline for generating agent-compatible tools in research domains. For scientific developers, such a pipeline enables the rapid creation of AI-usable tools, provided that APIs and their documentations are available in natural language. This improves development efficiency and allows developers to focus on their core research tasks. For end users, AI agents offer a natural language interface to interact with APIs, significantly reducing the learning curve and making tools more accessible. For the AI agent community, an automated pipeline allows for seamless integration of any available APIs, thereby greatly enhancing the versatility and capability of AI agents.

    
    


\begin{figure} 
    \centering
    \includegraphics[width=0.45\textwidth]{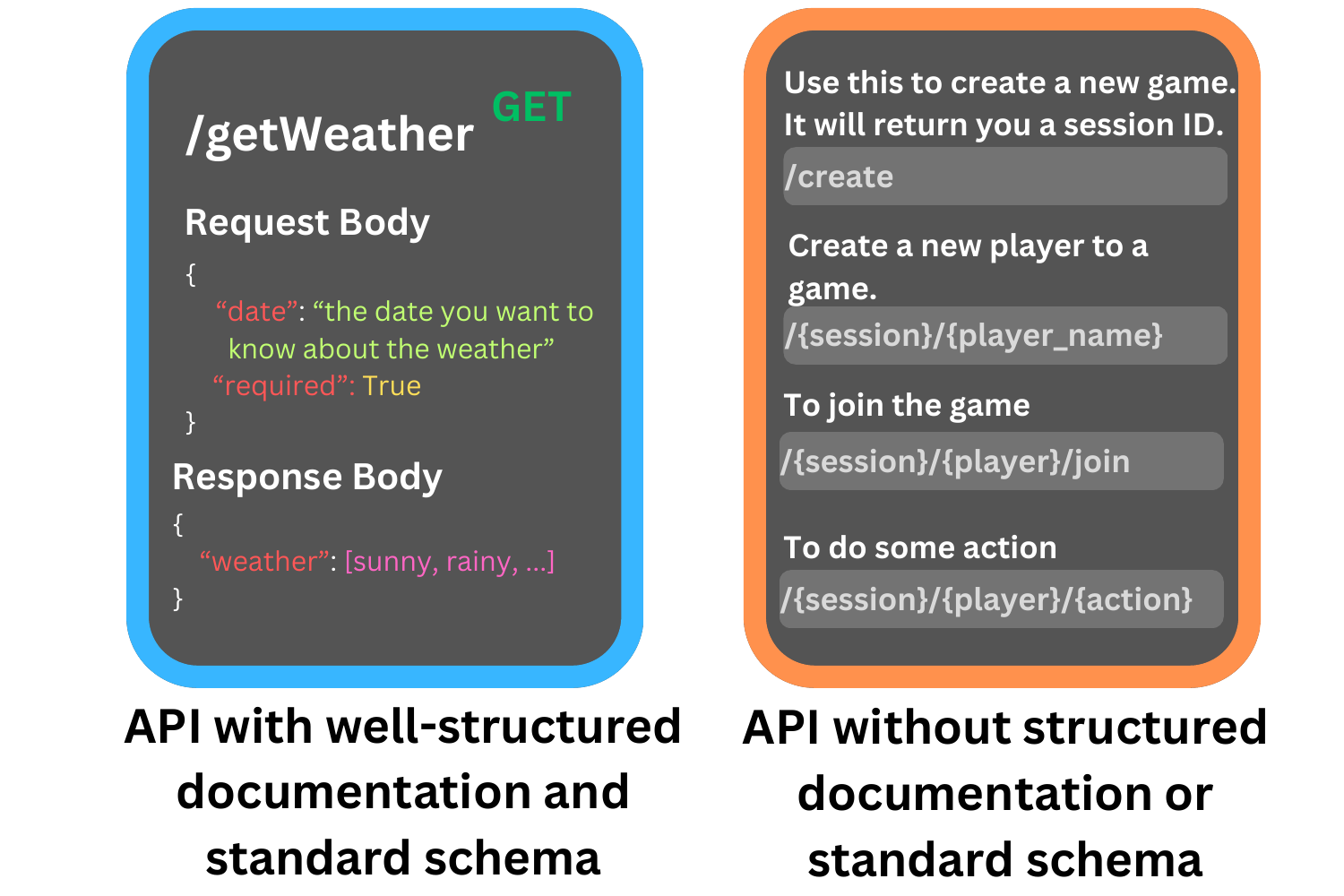}
    \captionsetup{width=0.9\linewidth}
    \caption{\textbf{The API Extraction Benchmark includes API documents with varying levels of structures.} For example, the API document example shown in the left follows a standardized schema. The one in the right is less structured as several fields are described in free-text. Our dataset prioritizes API variety and emphasizes the less structured cases. The diverse document structures in our dataset necessitate a general tool generation pipeline capable of processing various document formats effectively.}
    \label{fig:dataset}
\end{figure}

\begin{figure*}[htbp]
    \centering
    \includegraphics[width=\linewidth]{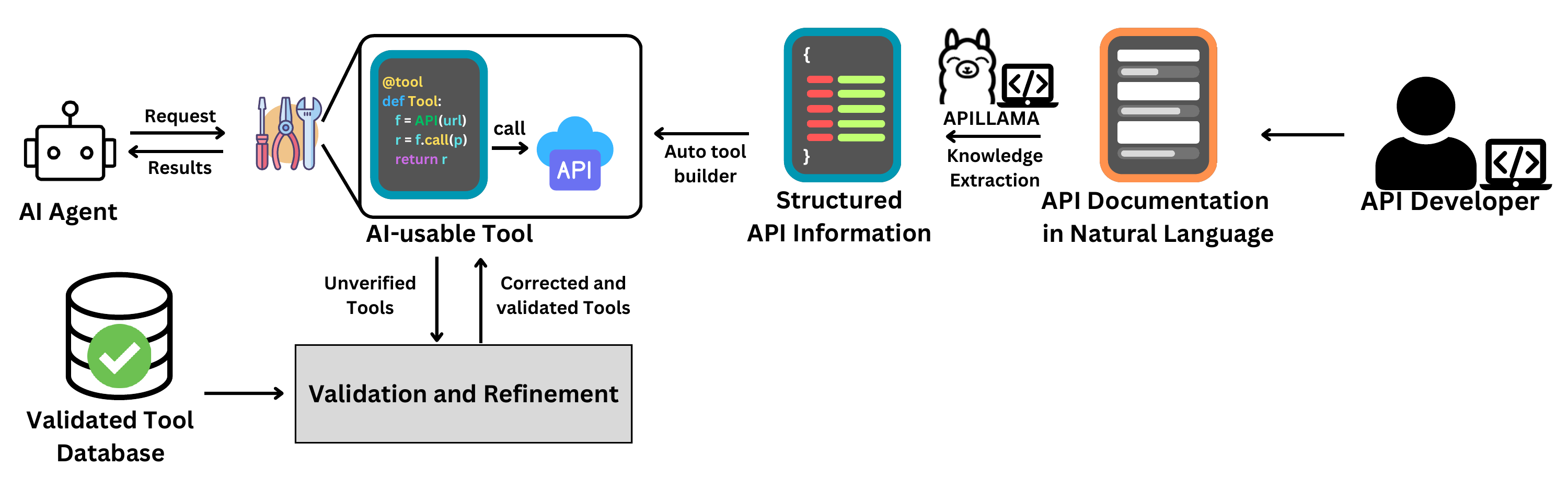}
    \caption{TooFactory: Automated pipeline for generating AI-usable tools from API documentations. The API documentations in free-text are processed by APILlama to extract structured information, which is used to build tools that interact with the corresponding APIs.}
    \label{fig:pipeline}
\end{figure*}

To address the limitations in tool agent development, we propose an open-source pipeline, \textbf{ToolFactory}\footnote{Code available at \href{https://github.com/coolkillercat/ToolFactory-review}{Link}}(Figure \ref{fig:pipeline}), which autonomously generates tools from any REST API documentation written in natural language, eliminating the need for human intervention. To demonstrate our approach, we also introduced an \textbf{API Extraction Benchmark} specifically constructed for automatic tool generation:
\begin{itemize}
    \item \textbf{Data collection:} We gathered 167 free API documentations from \href{https://apislist.com}{APIList.com}. Unlike previous benchmarks, we focused on APIs with diverse documentation formats (see Figure \ref{fig:dataset} for examples). This deliberate choice was made to ensure generalizability of our approach, as relying too heavily on APIs with uniform schemas (e.g., OpenAPI spec, RapidAPI schema) could limit the applicability of our approach. Documentation quality varies significantly among open-source APIs, which introduces challenges in understanding the APIs. We selected APIs that do not require authentication to simplify the evaluation framework and to avoid potential safety concerns. By using authentication-free APIs, we isolate the key challenge of extracting information from diverse documentation styles. In practical applications, authentication keys can be applied manually and incorporated into the AI-usable tools as part of their configuration.
    \item \textbf{Schema design:} Since the collected API documents are of various structures, we established a minimum required JSON schema(Appendix \ref{app:json-schema}) for standardizing the structured information that should be extracted from the documents, facilitating autonomous tool generation. Unlike common REST API schemas such as OpenAPI, our schema excludes the definition of API responses, as these are not always provided in documentation and do not directly impact API usage. In addition, we employ an LLM-based evaluator to handle different responses and classify possible errors(Section \ref{sec:validation}), as a complement to our schema design.
    \item \textbf{API document annotation:} We utilized GPT-4o in structured mode to produce standardized and well-organized API information for the API documents in our dataset. The quality of the data was examined and verified by human experts.
\end{itemize}

Our contributions can be summarized as follows:

\begin{itemize}
    \item Using the API extraction benchmark, we derived an open-source model, \textbf{APILlama}, by fine-tuning Llama 3~\cite{dubey2024llama3herdmodels} for the API extraction task using prompt tuning~\cite{lester2021powerscaleparameterefficientprompt}. Specifically, the model is trained to transform an API documentation into a JSON file that meets the predefined schema for the AI agent to perform as a tool, an example is provided in Appendix \ref{app:API-extraction-example}. We used prompt tuning with a soft prompt in the embedding layer to guide extraction. The soft prompt efficiently encodes the JSON schema and instruction(about 572 tokens) into shorter length(20 trainable virtual tokens). This approach minimizes trainable parameters, reduces overfitting, and preserves performance. Our experiment (Section \ref{sec:experiment-1}) shows that APILlama achieves significant improvement over the original LLaMA3 model, demonstrates comparable extraction capabilities to GPT-3.5's structured mode, and achieves higher accuracy in retrieving API parameters.
    \item  We export the converted extracted API information into Python functions as AI-usable tools that are compatible with LLM frameworks like LangChain~\cite{langchain}. To handle complex API responses, we developed an \textbf{evaluation pipeline} (Section \ref{sec:validation}) to verify the tools' functionality and diagnose errors when detected (Section \ref{sec:experiment-2}). We identified invalid parameter values as the primary cause of tool failures. To address this, we proposed a method leveraging a knowledge base to \textbf{improve parameter value inference} (Section \ref{sec:example-parameter-generation}).
    \item To further demonstrate the effectiveness of our approach, we used the generated tools to develop a Tool Agent to support glycomaterials research as a case study (Figure \ref{fig:glyco}). We collected relevant REST APIs contributed by researchers in the field and applied ToolFactory to them. To address the challenge of finding the proper parameter value such as database ID, we used information from verified tools to build a parameter database and refined other tools which have missing or imprecise information. As a result, we successfully \textbf{generated 92 verified tools} that enable access to and processing of glycan-related data, covering a wide range of tasks and glycan representations. Our findings also highlight the domain-agnostic nature of ToolFactory, as it can be effectively applied across diverse fields.
\end{itemize}


\section{Related Works}
\textbf{LLM for universal information extraction} Structured information extraction is important as it builds bridge between human-readable formats into machine-usable data. Lu et. al.\cite{lu2022unifiedstructuregenerationuniversal} proposed a universal text-to-structure framework for universal information extraction(UIE) using transformers, proving the potential of language models in encoding complex structures and producing structured outputs. Dagdelen et. al. \cite{dagdelen2024structured} finetuned LLAMA2 \cite{touvron2023llama2openfoundation} and GPT-3 \cite{brown2020languagemodelsfewshotlearners} to extract the desired information from scientific text that can be fit into defined JSON schema. Zhu et. al. \cite{zhu2024llmsknowledgegraphconstruction} proposed a multi-agent approach, encoporating GPT4 and GPT3.5 for automatic knowledge graph extraction and reasoning. OpenAI released GPT4o and GPT4o-mini structured mode, which supports generating structured outputs in an update in July \cite{OpenAIPydantic}. It forces GPT to follow the provided Pydantic structure or JSON schema. In this work, we employed both OpenAI model for data synthesis and finetuned LLAMA3 model for an opensource information extraction solution.

\textbf{Tool agents} LLM-based tool agents are able to reason through user queries, select and apply appropriate actions, and return the results of the chosen action. For example, Bran et. al. \cite{bran2023chemcrowaugmentinglargelanguagemodels} develop ChemCrow by integrating GPT-4 and 18 tools designed by experts. It was demonstrated that ChemCrow was able to effectively automate a variety of chemical tasks. Qin et. al. \cite{qin2023toolllmfacilitatinglargelanguage} collected 16k public REST APIs from the RapidAPI platform \cite{rapidapi}, and trained a tool retriever that can choose the most appropriate API in response to a user query. However, their tool-building approach relied on RapidAPI, limiting adaptability to other platforms without significant human effort. Similarly, RESTgpt\cite{song2023restgptconnectinglargelanguage} used OpenAPI Specification (OAS), restricting its applicability to APIs that provide OAS. Our goal is to create a universal tool-generation pipeline that works with any natural language API documentation.

\section{API Extraction Benchmark Dataset}
Previous research on tool agents, such as ToolLLM, has focused primarily on instructing LLM agents to use AI-usable tools and develop solution paths to answer user queries. Their AI-usable tools can be easily developed from APIs, supported by well-structured documentation, using relatively simple scripts. However, they did not focus on the challenges of developing AI-usable tools from APIs with less structured documentation. To fill this gap, we collected such APIs from \href{apilist.com}{APIList.com}. We introduce our dataset as follows:

\begin{itemize}

\item \textbf{Documentation styles} The API documentation we collected can be categorized into three levels based on the organization and clarity of the API descriptions: (1) \textbf{Fully organized} The documentation follows a well-defined template, providing all necessary information to call the API in a structured and comprehensive way. Use cases are clearly explained, often with example code. API documentation on platforms like RapidAPI Hub and Postman API typically fall into this category. (2) \textbf{Semi-organized} This type of documentation includes basic descriptions but lacks clarity for each endpoint. Some essential information may not be labeled with specific keywords and is instead embedded within general text. Additional effort is often required to identify key details. (3) \textbf{Unorganized} These documents are minimal, often missing example code or detailed descriptions. They require some level of inference and reasoning to understand the API's usage, with clues only available through endpoint names. We show that our benchmark consists of mostly semi-structure documentations, and only a few documentations are fully organized(Appendix \ref{app:api-doc-classification}).
\item \textbf{Selection criteria} We filtered for API documents that do not require API keys, allowing for easier and more convenient API access without needing authentication, which often involves submitting forms or linking payment methods. Although automating API key sign-up is feasible using web agents \cite{autosignup}, we opted for APIs without authentication requirements to streamline the process. In total, 347 unique API documents were selected and downloaded in HTML format. Since some links pointed to index pages or API information that was dynamically loaded via JavaScript, we employed a large language model to identify pages containing static HTML code with API endpoints. This approach ensured that we captured only the documentation with accessible and actionable API details.
\end{itemize}

Due to variations in the quality and completeness of API documentation, we extracted only the essential information needed for tool generation. Specifically, for each API documentation, we captured the \verb|base URL| and a list of endpoints. For each endpoint, we extracted the \verb|endpoint path|, \verb|required parameters|, \verb|optional parameters|, and a brief \verb|description|. In cases where the base URL was not specified, human annotation was necessary. The detailed schema used for this extraction is provided in the Appendix \ref{app:json-schema}.

To implement this extraction, we defined the schema using a Pydantic model and employed GPT-4o in structured mode to parse the HTML documents and extract the desired information. After filtering out pages that lack API information (primarily product index pages), we obtained 167 API documentation with 744 endpoints and extracted their structured information in JSON format. An example of input API documentation and output JSON structure is in Appendix \ref{app:API-extraction-example}.

\section{ToolFactory Pipeline}
\subsection{APILLAMA}
We fine-tuned the LLAMA3 model to develop an open-source solution for extracting structured information from API documentation. Recent LLM-based structured information extraction approaches rely on in-context learning, which involves providing a detailed JSON schema description or a one-shot input-output example. These contexts must be included in every inference, leading to a significant overhead of redundant input tokens. Our approach is straightforward and efficient—since we focus on extracting information based on a fixed schema(Appendix \ref{app:json-schema}), we employ soft prompt tuning to encode the task, compressing the JSON schema into fewer tokens(Appendix \ref{app:apillama-training}). This method significantly reduces token overhead during inference, as we compressed the original instruction which consisted of approximately 572 tokens (18 for the instruction and 554 for the JSON schema definition), into just 20 virtual tokens.

Despite its simplicity, our approach achieves competitive results. Even a fine-tuned 8B model demonstrates performance comparable to OpenAI’s GPT models(Section \ref{sec:experiment-1}). Prompt tuning also addresses the challenge of data availability; after filtering, the number of public, free APIs is relatively small, making smaller fine-tuned models less prone to overfitting.

We used 20 virtual tokens as the trainable prompt and applied 4-bit quantization to the main model, optimizing memory usage and training efficiency. We used Adam optimizer with learning rate 0.001. The training experiment was conducted on an NVIDIA A40 GPU and completed in 2 hours.

\subsubsection{Task Formalization}
APILLAMA aims to generate the structured annotation of an API from its documentation. Specifically, given the embedding of an API documentation $X\in \mathbb{R} ^{d \times n_X}$ that contains $n_X$ tokens with hidden dimension $d$, a pre-defined output schema $S$, a trainable instruction prompt $\varphi \in \mathbb{R} ^ {d \times n_I}$, a LLM $f_\theta$ with frozen weight $\theta$, the goal is to predict the corresponding structured information $Y_{(n_Y)}=(y_1,y_2,\dots,y_{n_Y})\in \mathbb{R} ^ {d \times n_Y}$. Basically, we want to maximize the probability of conditional generation $\text{P}_{\varphi}(\tilde{Y}|X;S;f_\theta)$. For a LLM $f_\theta$, we used the negative log likelihood loss for model training:
\begin{align*}
& \varphi = \argmin_{\varphi} - \frac{1}{n_Y} \sum_{i=1}^{n_Y} \mathcal{L}(y_i)  \\
& \mathcal{L}(y_i) = \log \text{P}_\varphi(y_i|Y_{(i-1)},X,S, f_{\theta})
\end{align*}
where $Y_{(i-1)}$ contains the ground truth tokens $y_1, y_2,\dots,y_{i-1}$. In testing, the ground truth will not be provided to the model. Instead, we use an auto-regressive method to predict the tokens:
\begin{align*}
& \tilde{y}_i = \argmax_{y_i} \text{P}_\varphi(y_i|\tilde{Y}_{(i-1)}, X,S,f_\theta) \\
\end{align*}
$\tilde{Y}_{(i-1)}$ represents all previously predicted tokens. During implementation, the model generates a special \verb|<EOS>| token to signal the end of generation, which is masked and excluded from attention calculations.

\subsubsection{Evaluation of Structured Information Extraction} \label{sec:evaluation-SIE}
While we trained the model as a text generation task, our ultimate goal is to extract the proper knowledge from any given API documentation. The token level prediction alone cannot reveal the performance of the model. Hence, we evaluated several aspects of the generated JSON files:
\begin{itemize}
    \item \textbf{Structure Correctness:} We first validate if the generated JSON file matches with our predefined schema. The \textbf{Valid Ratio} metric evaluates the model's overall ability to generate correctly structured JSON files. 
    
    \item \textbf{Semantic Similarity:} The name and description for the API endpoint and API parameters are crucial for enabling the AI agent to understand how to use the tool correctly and effectively. To evaluate the model's ability to extract precise semantic information, we embed the endpoint name, endpoint description and parameter description using \textit{st-codesearch-distilroberta-base}\cite{st-codesearch-distilroberta-base}, a sentence transformer trained on \textit{CodeSearchNet}\cite{husain2019codesearchnet}. This embedding model is designed to handle both natural language and coding-related information. We then calculate the cosine similarity between the prediction and ground-truth embeddings for three metrics: \textbf{Endpoint Name Similarity}, \textbf{Endpoint Description Similarity}, and \textbf{Parameter Description Similarity}.  
    
    \item \textbf{Functional Parameter Accuracy:} The API server validates the payload and processes parameters based on JSON keys. An invalid parameter name can fail an API call. Additionally, API methods (e.g., GET, POST) and parameter types (e.g., string, int) are critical components that can be categorized. We therefore measure \textbf{Parameter Precision and Recall}, \textbf{Method Accuracy}, and \textbf{Parameter Type Accuracy}. 
\end{itemize}

The experiment results (Section \ref{sec:experiment-1}) show that APILLAMA performed nearly perfect (97\%) in generating correct structures and is comparable with proprietary models in semantic similarity and value matching.

\subsection{Tool-generation Process}
Once the necessary API information is extracted, the extracted JSON files are processed by scripts to generate AI-usable tools. These tools are designed to adapt the API's functionality into formats compatible with popular agent-building frameworks. We support exporting tool functions as Python functions, which can be directly integrated into open-source frameworks like LangChain\cite{langchain} and OpenAgents\cite{xie2023openagentsopenplatformlanguage}. Additionally, tools can be exported as OpenAPI YAML files, enabling seamless import into systems like OpenAI GPTs \cite{openaigpts}. The details of the tool generation algorithm, including format adaptation and compatibility handling, are provided in Appendix \ref{app:tool-generator}.

\subsection{Tool Validation} \label{sec:validation}
An evaluation process is employed to validate each generated tool. Only tools that pass this evaluation will be made available for AI agents. We test a tool using example parameter values and consider it valid only if: (1) it returns a status code of 200, and (2) its content is neither empty nor an error message, as verified by an GPT-4o based evaluator. Otherwise, the tool is considered to have failed validation. We categorize the tool errors into 6 types: Missing Endpoint Path, Missing Base URL, Failed Validation, Abnormal Response, No Parameter Value and Wrong Parameter Value. A detailed explanation can be referred in Appendix \ref{app:error-validation}

In addition, we categorize error causes into four main categories: \textbf{C1} Missing API Documentation Details, \textbf{C2} Incorrectly Extracted URL Path, \textbf{C3} Incorrect Parameter Values, and \textbf{C4} Server-Side Errors. For each category, we provide a range of possible error diagnosis, from the most conservative to the most aggressive (see Appendix \ref{app:tab:error-estimation}). We find that most of the errors are caused by \textbf{C3}, which inspires us to develop an better approach to produce high-quality parameter values.



\subsection{Parameter Value Inference}
\label{sec:example-parameter-generation}
\begin{figure}[htbp]
\centering
\includegraphics[width=\linewidth]{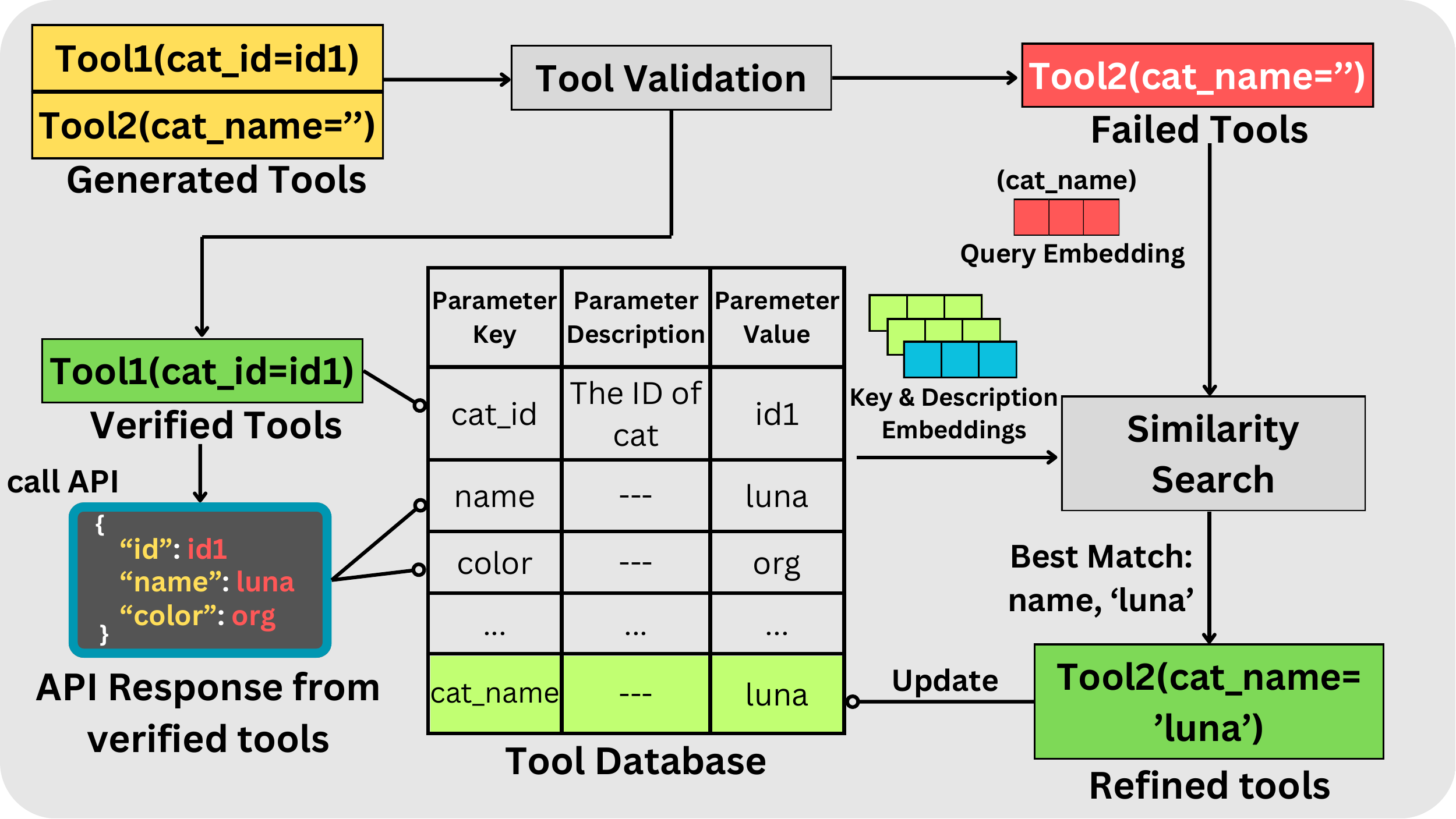}
\caption{A parameter database is constructed using validated tools, enabling parameter value inference based on the semantic similarity of parameter keys and descriptions.}
\label{fig:parameter-value-inference}
\end{figure}
API documentations vary in quality, and parameter value specifications are often missing, especially in poorly documented APIs. To tackle this problem, we propose the following method to infer missing parameter value specifications of an API from validated tools in the same application domain. 

First, we built a knowledge base about parameters from validated tools. Each parameter entry contains an embedding vector indicating its semantics. Such knowledge can be extracted from the corresponding API documentations. Alternatively, we can extract such knowledge from the JSON responses of successfully executed tools. If a parameter, say $\gamma$ is associated with a description (but not sufficient to determine its value), we embed its name and description separately. Using these embedding vectors, we retrieve the top 10 most similar parameters from the knowledge base—5 based on description cosine similarity and 5 based on parameter key similarity. Then we apply the value of each retrieved parameter to test the tool. If a candidate value passes the validation, we will use it to annotate $\gamma$, and save it to the parameter knowledge base. Figure \ref{fig:parameter-value-inference} demonstrates our approach.



\begin{table*}[htbp]
\centering
\resizebox{\linewidth}{!}{%
\begin{tabular}{lccccccccc}
\toprule
  \multicolumn{1}{c}{}  & \multicolumn{1}{c}{JSON Schema} & \multicolumn{3}{c}{API Endpoint}                                                                                                                                                   & Method   & \multicolumn{4}{c}{Parameter}                                                                                                                                                                                \\ \cmidrule(rl){2-2} \cmidrule(rl){3-5} \cmidrule(rl){6-6} \cmidrule(rl){7-10}
Model     & Valid Ratio & \multicolumn{1}{c}{\# Matched} & \multicolumn{1}{c}{\begin{tabular}[c]{@{}c@{}}Name\\ Similarity\end{tabular}} & \begin{tabular}[c]{@{}c@{}}Description\\ Similarity\end{tabular} & Accuracy & \multicolumn{1}{c}{Precison} & \multicolumn{1}{c}{Recall} & \multicolumn{1}{c}{\begin{tabular}[c]{@{}c@{}}Description\\ Similarity\end{tabular}} & \begin{tabular}[c]{@{}c@{}}Type\\ Accuracy\end{tabular} \\ \midrule

LLaMA3 & 0 & 0 & 0 & 0 & 0 & 0 & 0 & 0 & 0 \\
LLaMA3+one shot                  & 0.59        & \multicolumn{1}{c}{31}         & \multicolumn{1}{c}{\textbf{0.98}}                                                      & 0.88                                                             & 0.98    & \multicolumn{1}{c}{0.84}     & \multicolumn{1}{c}{0.82}   & \multicolumn{1}{c}{0.88}                                                             & \textbf{1.00}                                                    \\ 
GPT3.5+one shot                  & 0.65        & \multicolumn{1}{c}{44}         & \multicolumn{1}{c}{0.90}                                                      & 0.87                                                             & 0.98     & \multicolumn{1}{c}{0.88}     & \multicolumn{1}{c}{0.87}   & \multicolumn{1}{c}{\textbf{0.93}}                                                             & \textbf{1.00}                                                    \\ 
GPT3.5 structured mode & \textbf{1.00}        & \multicolumn{1}{c}{\textbf{84}}         & \multicolumn{1}{c}{0.96}                                                      & 0.80                                                             & 0.97     & \multicolumn{1}{c}{0.83}     & \multicolumn{1}{c}{0.82}   & \multicolumn{1}{c}{\textbf{0.93}}                                                             & 0.63                                                    \\ 
\textbf{APILLAMA(Ours)}                         & 0.97        & \multicolumn{1}{c}{66}         & \multicolumn{1}{c}{0.94}                                                      & \textbf{0.89}                                                             & \textbf{1.00}     & \multicolumn{1}{c}{\textbf{0.92}}     & \multicolumn{1}{c}{\textbf{0.92}}   & \multicolumn{1}{c}{0.91}                                                             & \textbf{1.00}                                                    \\ \bottomrule
\end{tabular}%
}
\caption{A comparison of API information extraction result. Larger metric values indicate better performance. Metric explanation can be referred at Section \ref{sec:evaluation-SIE}}
\label{tab:extraction-experiment}
\end{table*}

\section{Results}
Using the API Extraction Benchmark, we evaluated the capability of APILLAMA to extract API information from documentations in natural language and tested the automated tool generation process. We also demonstrated a case study where we built an AI-agent using the tools automatically generated by APILLAMA.

\subsection{Structured API Knowledge Extraction via APILLAMA}
\label{sec:experiment-1}
We investigated the feasibility of combining LLM and soft prompt for extracting API knowledge.

\textbf{Model} APILLAMA consists of the LLAMA3-8B-instruct model, which is frozen, and 20 trainable soft prompt tokens. In addition, we extend the token limit from 8,192 to 10,240 to handle longer documents. 

\textbf{Training setting} The input of the model is the text in an API documentation, and the output is the corresponding structured information in JSON. We randomly split the API Extraction Benchmark into 80\% for training and 20\% for testing. The test set contains 34 API documents and 117 different endpoints. More setting details are provided in Appendix \ref{app:apillama-training}.

\textbf{Metrics} See Section \ref{sec:evaluation-SIE}. 

\textbf{Baselines} We chose the original LLaMA3-8B-instruct and GPT3.5 as the baseline models. We tested how the supporting information will affect the model performance. In the "one-shot learning" setting, we include an example in the input prompt that consists of an API document as the input along with its corresponding JSON annotation as the target. On the other hand, we defined the target JSON schema with Pydantic\cite{pydantic} package, which can be used by the function calling mode\cite{openaifunctioncalling} of GPT-3.5. This guarantees GPT-3.5 model to understand and produce the desired JSON structure.

The results are shown in Table \ref{tab:extraction-experiment}. APILLAMA demonstrates strong performance, achieving competitive or superior scores across most metrics. In contrast, the vanilla LLaMA3 model performs poorly even when provided with a one-shot example. APILLAMA significantly improves the retrieval of API endpoints and the generation of accurate API annotations. Compared to GPT-3.5 in structured mode, APILLAMA retrieves fewer endpoints but excels in accurately extracting endpoint descriptions and parameters. The limited number of retrieved endpoints is likely due to the reasoning capability of the base LLaMA3-8B model, as its weights were not modified during fine-tuning. We attribute APILLAMA’s performance improvements to two key factors: (a) The soft prompt effectively describes the structured information extraction task, allowing the model to better understand the schema; (b) By minimizing the number of tokens used for instructions, APILLAMA enables its base model to focus more on the actual data within the prompt.

\subsection{Automatic Tool Generation} \label{sec:experiment-2}
We evaluated the performance of ToolFactory in generating AI-usable tools. These tools were derived from the structured API information extracted in Section \ref{sec:experiment-1} and tested following the procedure described in Section \ref{sec:validation}. Each tool was implemented as a Python function that takes an API endpoint and its corresponding parameters as inputs, processes the request, and returns the API response. An example of such a tool is provided in Appendix \ref{app:tool-example}.

The results are summarized in Table \ref{tab:experiment-2}. While APILLAMA extracted fewer endpoints—likely due to the reasoning limitations of its base model—it produced the highest number of tools that passed validation. In the API Extraction Benchmark, 50 tools generated from ground-truth JSON files passed the validation test. APILLAMA was able to generate 26 tools (successful rate of 52\%).

We also analyzed the errors (Appendix Table \ref{app:tab:tool-errors}) in the generated tools. The primary cause of failures in APILLAMA and GPT-3.5 structured mode stemmed from incorrect parameter values. Notably, 30\% to 55\% of the tools generated by APILLAMA failed due to improper parameter values. This analysis highlights a key limitation of the current LLM-based approach and motivates us to incorporate domain knowledge to augment parameter values in the next experiment.

\begin{table}[htbp]
\resizebox{\linewidth}{!}{
\begin{tabular}{lcc}
\toprule
\multicolumn{1}{c}{}                & \multicolumn{1}{l}{\textbf{\# Extracted tools}} & \multicolumn{1}{l}{\textbf{\# Passed Tools}} \\
\midrule
LLaMA3+one shot & 31                                  & 3                                \\
GPT3.5+one shot                     & 44                                  & 19                               \\
GPT3.5 structured mode              & 84                                  & 25                               \\
\textbf{APILLAMA(Ours)}             & 66                                  & \textbf{26}                     \\ \midrule
API Extraction Benchmark    & 117                                 & 50                               \\ \bottomrule
\end{tabular}
}
\caption{Result of tool validation}
\label{tab:experiment-2}
\end{table}

\subsection{Case Study: Automated Tool Agent Generation for Glyco Research} \label{glycogpt}

\begin{figure}[h]
    \centering
    \includegraphics[width=0.7\linewidth]{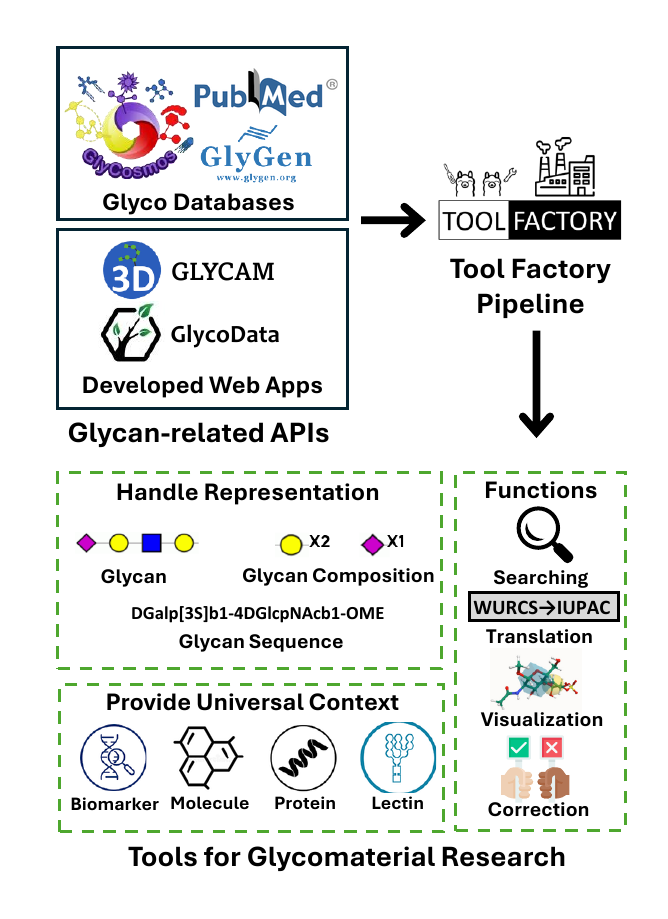}
    \caption{\textbf{AI Agent for Glycomaterial Research with Automated Tool Generation} \quad By automating tool generation, the AI agent simplifies database access and supports glycan-related tasks such as searching, drawing, and format conversion. ToolFactory generated 92 validated AI-usable tools for various tasks across several databases. A web demo is developed using the OpenAgents framework.}
    \label{fig:glyco}
\end{figure}

We applied ToolFactory to generate tools to support Glyco research and built an autonomous tool agent as the proof-of-principle. We collected the REST APIs of the most frequently used glyco databases, including \textbf{GlycoData}~\cite{glylcodata}, \textbf{GlyGen}~\cite{york2020glygen}, \textbf{GlyTouCan}~\cite{tiemeyer2017glytoucan}, \textbf{KEGG GLYCAN}~\cite{hashimoto2006kegg}, \textbf{Glycosmos}~\cite{yamada2020glycosmos}, \textbf{Glyconnect}~\cite{alocci2018glyconnect}, \textbf{The O-GlcNAc Database}~\cite{wulff2021human}, \textbf{GLYCAM}~\cite{Glycam}, \textbf{Protein API}~\cite{10.1093/nar/gkx237}, \textbf{PubChem}~\cite{kim2016pubchem} and \textbf{UniLectin}~\cite{imberty2021unilectin}. In total, 92 automatically generated tools passed validation, their functionalities cover data searching, calculation, format translation, correction, and visualization (Figure \ref{fig:glyco}). 

\subsubsection{Contribution to Glyco Research Community} 

Our AI agent provides natural language interfaces for researchers to access web services (e.g., databases and utility APIs) without requiring technical expertise, which greatly facilitates the usage of online resources and benefits researchers in several ways:

\textbf{(1) Automatic cross-database integration\quad} No single database fulfills all information needs due to their specific focuses. For example, GlyTouCan catalogs glycan structures, KEGG GLYCAN maps pathways and reactions, and PubChem offers general molecular data. Our AI agent integrates these diverse sources for seamless information access. 

\textbf{(2) Tool synergy\quad} Individual tools are often limited to specific scenarios, but our AI agent enhances their applicability by integrating them into cohesive workflows, including ID conversion, database querying, data normalization, visualization, and resolution of inconsistencies. Appendix Table \ref{app:tab:glycanid} shows an example of a glycan being represented using various formats. This integration broadens the applicability of certain tools. For instance, the GLYCAM 3D visualization tool, initially limited to GLYCAM strings, now supports multiple glycan formats through automated conversion.



\textbf{(3) Facilitate the development and adoption of new services\quad} Using ToolFactory, our AI agent can easily adopt new APIs and datasets, keeping its tools up-to-date. This enables researchers to prioritize scientific exploration over technical integration. In addition, the AI agent can serve as a platform for developing, sharing, and publishing applications, fostering dissemination and collaboration.


In the following, we highlight the technical details for solving the challenge of insufficient domain knowledge, which is common when developing domain-specific agent. For example, databases may use different reference IDs (see Appendix Table \ref{app:tab:glycodataset} for an example), requiring the agent to distinguish and pick the correct ID carefully. However, the information may not be covered in the API documentation, or is simply referred to as "ID" without any further information. The parameter value inference method proposed in Section \ref{sec:example-parameter-generation} was deployed to address this problem, and its effectiveness is evaluated below.

\textbf{Setting} We used 92 validated tools in this experiment. We first built a parameter database utilizing the parameter examples of verified documents and response JSONs. For each parameter name-value pair, we added the information of its original API documentation to track the source. To simulate a scenario where the information about setting the value of a parameter is missing, we deliberately removed the parameter value information to test the inference method. We also restricted the access to any other information in the source document of that parameter. Each method will generate candidate settings for the parameters. An inference result is deemed successful if it allows the tool to pass the validation.

\textbf{Our parameter inference method} We retrieve five parameters which have the most similar description semantic similarity and another five parameters with highest parameter key similarity, resulting in at most 10 candidate parameters for each query parameter. We filter out candidates with similarity smaller than 0.5. If an API has multiple parameters, we first generate candidate values for each parameter, and sample the best 20 value combinations to test the tool.

\textbf{Parameter inference using GPT-4o} We prompt GPT-4o to predict candidate parameters. To ensure fair comparison, we allow GPT-4o to make up to 10 candidate guesses per tool, as our approach requests 10.06 times per tool on average. The failed guesses will be provided in the prompt, ensuring GPT-4o to generate diverse values.

\textbf{Results} Using the leave-one-API-out test setting, our approach successfully infers parameter values for 33 tools. Meanwhile, GPT-4o managed to find the parameter values for only 17 tools. The results clearly indicate the superiority of our approach over GPT-4o.

\section{Conclusion}
In this work, we designed a pipeline, \textbf{ToolFactory}, that can automatically convert natural language REST API documentation into tools usable by AI agents. This pipeline not only translates natural language into agent-compatible tools but also diagnoses and optimizes non-standard or inaccurate information in REST API documentation. We first built the API Extraction Benchmark, which collects API documentation of varying quality and standardizes it into JSON format. Next, we trained APILLAMA using prompt tuning, enabling it to extract structured information from the documentation effectively. Our experimental results demonstrate that APILLAMA achieves strong performance in information extraction, with the number of tools generated being comparable to GPT-based models. We identified inaccurate parameter values as a key factor affecting APILLAMA’s performance and proposed a method to enhance parameter quality by constructing a domain-specific knowledge base. Finally, we applied ToolFactory to glycomaterial research and successfully developed an AI agent. Our work facilitates the creation of domain-specific agents, helping developers reduce the development and learning costs associated with these APIs.

\bibliographystyle{ieeetr}
\bibliography{ijcai25}

\newpage
\appendix

\renewcommand{\thetable}{\Alph{section}\arabic{table}}
\setcounter{table}{0}
\renewcommand\thefigure{\thesection\arabic{figure}}    
\setcounter{figure}{0}
\onecolumn

\lstdefinelanguage{json}{
    basicstyle=\normalfont\ttfamily,
    numbers=left,
    numberstyle=\scriptsize,
    stepnumber=1,
    numbersep=8pt,
    showstringspaces=false,
    breaklines=true,
    frame=lines,
    backgroundcolor=\color{background},
    literate=
     *{0}{{{\color{numb}0}}}{1}
      {1}{{{\color{numb}1}}}{1}
      {2}{{{\color{numb}2}}}{1}
      {3}{{{\color{numb}3}}}{1}
      {4}{{{\color{numb}4}}}{1}
      {5}{{{\color{numb}5}}}{1}
      {6}{{{\color{numb}6}}}{1}
      {7}{{{\color{numb}7}}}{1}
      {8}{{{\color{numb}8}}}{1}
      {9}{{{\color{numb}9}}}{1}
      {:}{{{\color{punct}{:}}}}{1}
      {,}{{{\color{punct}{,}}}}{1}
      {\{}{{{\color{delim}{\{}}}}{1}
      {\}}{{{\color{delim}{\}}}}}{1}
      {[}{{{\color{delim}{[}}}}{1}
      {]}{{{\color{delim}{]}}}}{1},
}

\section{API-extraction Schema} \label{app:json-schema}
We defined the JSON schema to extract API information. In pratice, this is achieved by defining the pydantic object of API objects.
\begin{lstlisting}[language=json,firstnumber=1]
{
  "$schema": "http://json-schema.org/draft-07/schema#",
  "definitions": {
    "Parameters": {
      "type": "object",
      "properties": {
        "name": {
          "type": "string",
          "description": "Name of the parameter"
        },
        "type": {
          "type": "string",
          "description": "Type of the parameter"
        },
        "description": {
          "type": "string",
          "description": "Description of the parameter. If the parameter is categorical, please list all possible values."
        },
        "default": {
          "description": "Default value of the parameter"
        },
        "example": {
          "description": "Example value of the parameter"
        }
      },
      "required": ["name"]
    },
    "Endpoint": {
      "type": "object",
      "properties": {
        "name": {
          "type": "string",
          "description": "Name of the endpoint"
        },
        "description": {
          "type": "string",
          "description": "Description of the endpoint"
        },
        "method": {
          "type": "string",
          "description": "Method of the endpoint"
        },
        "url": {
          "oneOf": [
            { "type": "string" },
            { "type": "array", "items": { "type": "string" } }
          ],
          "description": "URL of the endpoint, start with http:// or https://"
        },
        "headers": {
          "type": "array",
          "items": { "type": "string" },
          "description": "Headers of the endpoint",
          "default": []
        },
        "required_parameters": {
          "type": "array",
          "items": { "$ref": "#/definitions/Parameters" }
        },
        "optional_parameters": {
          "type": "array",
          "items": { "$ref": "#/definitions/Parameters" }
        }
      },
      "required": ["name", "method", "url"]
    },
    "API": {
      "type": "object",
      "properties": {
        "title": {
          "type": "string",
          "description": "Title of the API"
        },
        "endpoints": {
          "type": "array",
          "items": { "$ref": "#/definitions/Endpoint" }
        }
      },
      "required": ["endpoints"]
    }
  },
  "type": "object",
  "properties": {
    "API": { "$ref": "#/definitions/API" }
  }
}
\end{lstlisting}

\newpage

\section{API Extraction Benchmark}
\subsection{API Documentation Classification} \label{app:api-doc-classification}
\begin{figure}[htbp]
\centering
\subfigure[]{
\includegraphics[width=0.3\linewidth]{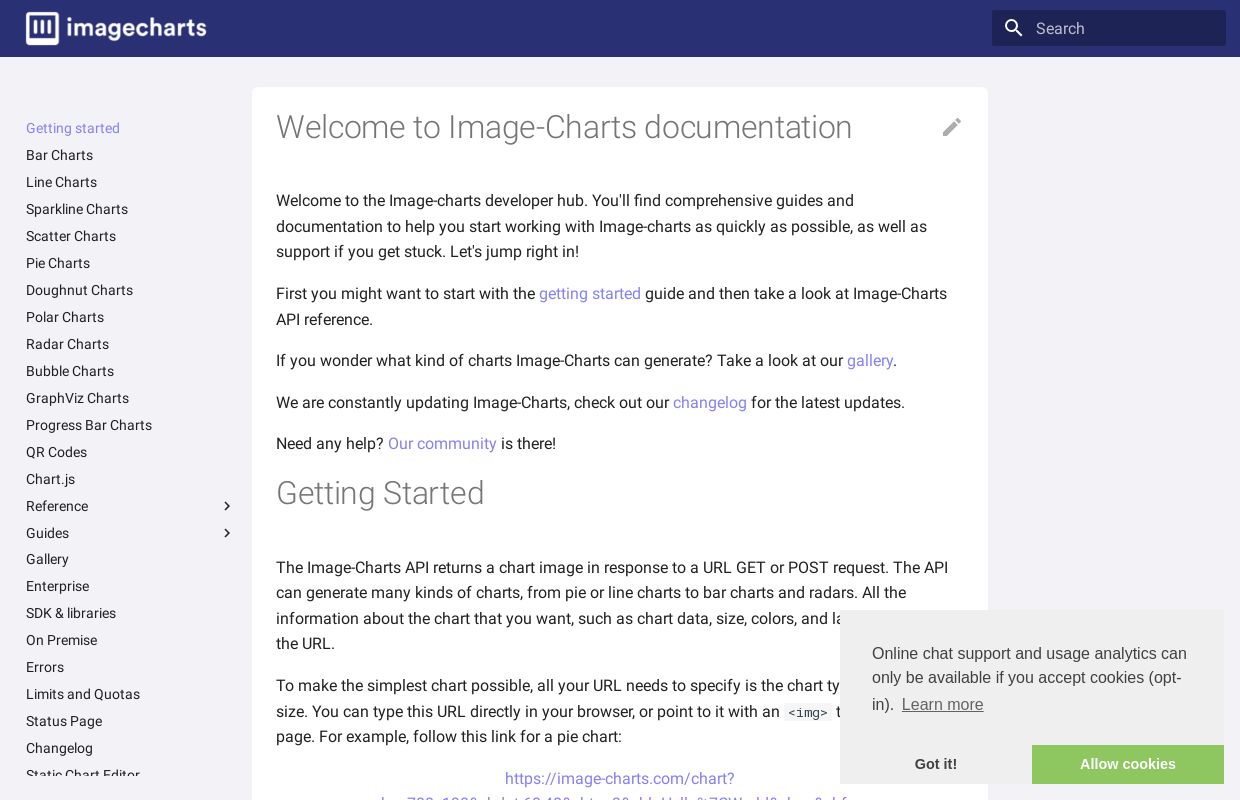}
}
\subfigure[]{
\includegraphics[width=0.3\linewidth]{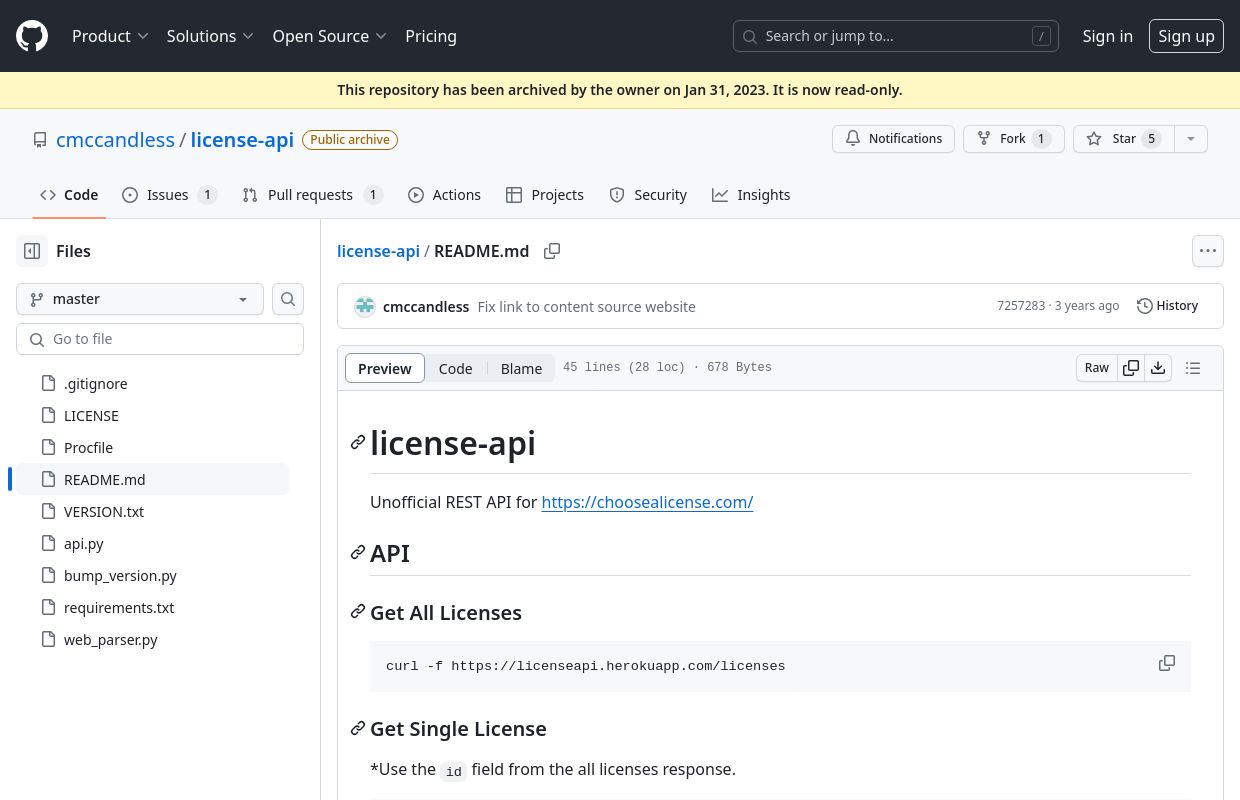}
}
\subfigure[]{
\includegraphics[width=0.3\linewidth]{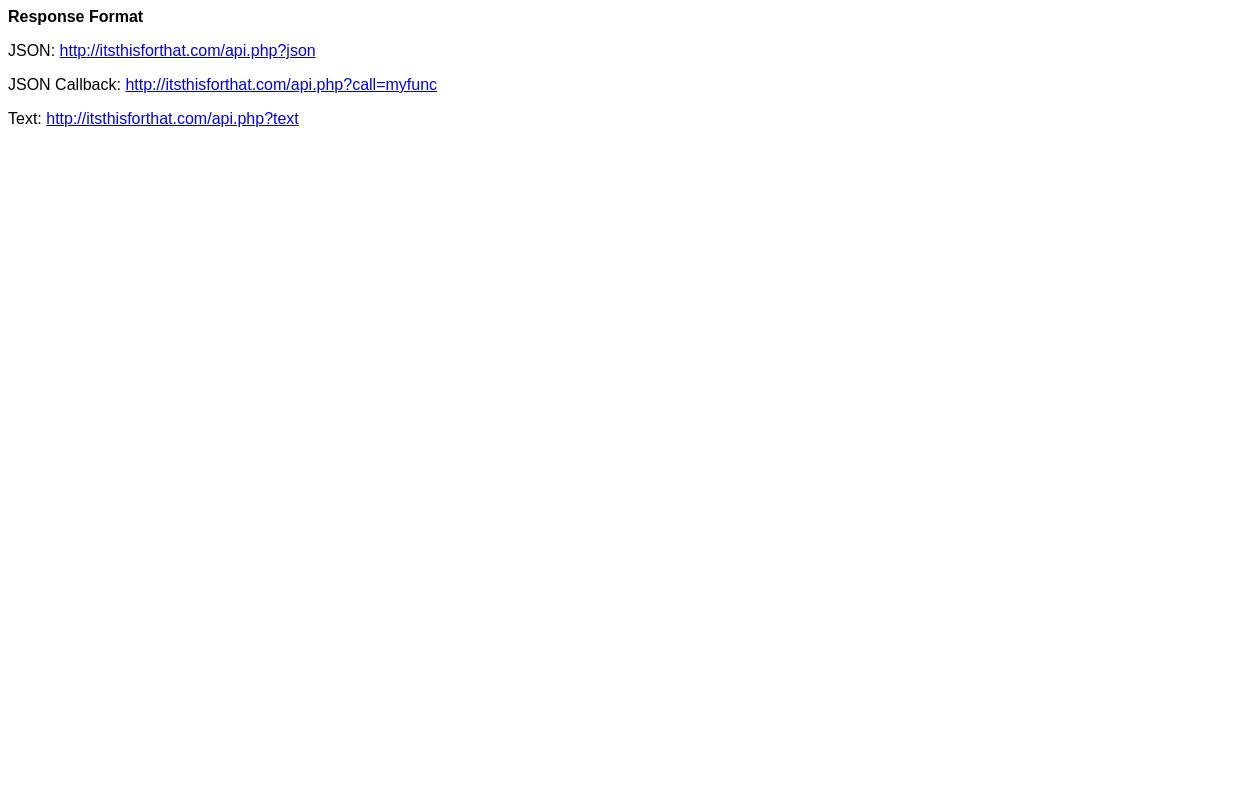}
}
\caption{Example of API documentation of each category. (left) is an example of organized API documentation \href{https://documentation.image-charts.com/?utm_source=apislist.com}{\textbf{ link}}}, (mid) is the example of semi-organized documentation \href{https://github.com/cmccandless/license-api/blob/master/README.md?utm_source=apislist.com}{\textbf{ link}}, and (right) is the unorganized API documentation \href{https://itsthisforthat.com/api.php?utm_source=apislist.com}{\textbf{ link}}
\label{app:fig:api-webpage-example}
\end{figure}

We classify API documentations into 3 categories: \textbf{Fully organized}, \textbf{Semi-organized} and \textbf{Unorganized}, based on their clarity and completeness of information. Using GPT-4o with chain-of-thought reasoning (Appendix \ref{app:doc-classification-prompt}), we categorized all API documentation in the API Extraction Benchmark. Out of the total, 24 documents were classified as Fully Organized, 134 as Semi-Organized, and 9 as Unorganized. Figure \ref{app:fig:api-webpage-example} provides examples of API documentation in each category. Notably, none of the documentation is fully standardized, and Semi-Organized and Unorganized documents frequently lack clarity or essential API details.

\subsection{Extraction Task Example}
\label{app:API-extraction-example}
In API Extraction Benchmark, the input is the text from \href{https://pokemontcg.io/?ref=apilist.fun}{the API documentation(HTML) page}, while the output is a JSON string following our API-extraction Schema. Below is the input example:

\noindent\fbox{%
    \parbox{\textwidth}{%
PokÃ©mon TCG Developers PokÃ©mon TCG Developers JOIN THE COMMUNITY OF DEVELOPERS BUILDING WITH THE POKÃ\\x89MON TCG API Next The most comprehensive dataset you can find From Base Set through Sword \& Shield , the PokÃ©mon TCG API has the card data you need to build the ultimate application, whether that be mobile or web. Get access to prices, hi-res images, and more. Focus on your application and let us worry about the card data. Some features of the PokÃ©mon TCG API include: Prices by TCGplayer Hi-Res Images Detailed Search Queries Developer SDKs With a large community contributing to the project through various SDKs, development for you and your team has never been easier. In-depth details of every card All data delivered in JSON format such as types, attacks, abilities, and more. For Developers The PokÃ©mon TCG API allows developers to ingest card data programatically. The API exposes card and set details in an easy to consume format. Stay up to date with new sets, and access new data right away on launch day. Advanced Search Queries Perform advanced search queries with a familiar Lucene-like syntax. Find cards by name, type, release date, legality, and more. Use multiple types of operators to filter to specific cards. This API supports operators such as logical AND, OR, NOT, and range operators. View the documentation for more details. GET https://api.pokemontcg.io/v2/cards?q=name:gardevoir (subtypes:mega OR subtypes:vmax) Gain access to the PokÃ©mon TCG Developer Portal to start using the PokÃ©mon TCG API The PokÃ©mon TCG Developer Portal lets you manage your account and API Key associated with the PokÃ©mon TCG API. Creating an account will get you access to higher rate limits and no IP restrictions. Get Started Questions? We are here to help you. If you have general usage questions, it is highly recommended to ask your questions at community Discord channel. There are tons of devs who are more than willing to help you out with any questions regarding the API. If you are an individual or business who is interested in receiving higher rate limits for the API, please contact Andrew via email: [email\\xa0protected] . I try to respond to all inquiries within 72 hours. API Documentation PokÃ©mon TCG API Documentation PokÃ©mon Card Search Engine PokÃ©mon TCG Guru Donate Patreon , Kofi Discord https://discord.gg/dpsTCvg Email [email\\xa0protected] Elsewhere GitHub Twitter LinkedIn This website is not produced, endorsed, supported, or affiliated with Nintendo or The PokÃ©mon Company.
}
}

And the output example:

\begin{lstlisting}[language=json]
    
{
    "title": "Pok\\u00e9mon TCG API Documentation",
    "endpoints": [
        {
            "name": "Search Cards",
            "description": "Perform advanced search queries to find cards by name, type, release date, legality, and more.",
            "method": "GET",
            "url": "https://api.pokemontcg.io/v2/cards",
            "headers": [],
            "required_parameters": [
                {
                    "name": "q",
                    "type": "string",
                    "description": "The search query using Lucene-like syntax.",
                    "default": null,
                    "example": "name:gardevoir"
                }
            ],
            "optional_parameters": []
        }
    ]
}
\end{lstlisting}

\section{JSON-To-Tool generation}
\label{app:tool-generator}
JSON-To-Tool generation is mostly about engineering. As the documentations are written in various formats, multiple conditions need to be considered, especially when handling URLs. It is common for the API server to only support a specific input pattern, which is not mentioned in the documentation or the error message. To make the autogenerated tool more robust, we considered the following procedures in our tool generator:
\begin{itemize}
    \item \textbf{parameter handling} Documentations use various ways to represent path parameters. We set matching rules to find commonly used patterns such as "\verb|:param|", "\verb|{param}|", "\verb|<param>|" etc.
    \item \textbf{encoding correction} To pass some special characters such as "+" or "=", the URL will use an encoding method known as percent encoding. Usually this won't be an issue, but for some APIs this need to be done before passing the parameters.
    \item \textbf{required parameter checking} As APIs may not necessarily return the error information, we added a required parameter validation step in the generated tool so that we can report any missing parameter errors to the AI agent.
    
\end{itemize}

\section{Tool Example} \label{app:tool-example}
By appling the JSON-To-Tool generation script, we can convert the structured API information into an excutable tool. In this work, a tool is a python function which can be called by the LLM. A typical tool example for 'Pokemon TCG API' is shown below.

\definecolor{maroon}{cmyk}{0, 0.87, 0.68, 0.32}
\definecolor{halfgray}{gray}{0.55}
\definecolor{ipython_frame}{RGB}{207, 207, 207}
\definecolor{ipython_bg}{RGB}{247, 247, 247}
\definecolor{ipython_red}{RGB}{186, 33, 33}
\definecolor{ipython_green}{RGB}{0, 128, 0}
\definecolor{ipython_cyan}{RGB}{64, 128, 128}
\definecolor{ipython_purple}{RGB}{170, 34, 255}

\lstset{
    breaklines=true,
    extendedchars=true,
    literate=
    {á}{{\'a}}1 {é}{{\'e}}1 {í}{{\'i}}1 {ó}{{\'o}}1 {ú}{{\'u}}1
    {Á}{{\'A}}1 {É}{{\'E}}1 {Í}{{\'I}}1 {Ó}{{\'O}}1 {Ú}{{\'U}}1
    {à}{{\`a}}1 {è}{{\`e}}1 {ì}{{\`i}}1 {ò}{{\`o}}1 {ù}{{\`u}}1
    {À}{{\`A}}1 {È}{{\'E}}1 {Ì}{{\`I}}1 {Ò}{{\`O}}1 {Ù}{{\`U}}1
    {ä}{{\"a}}1 {ë}{{\"e}}1 {ï}{{\"i}}1 {ö}{{\"o}}1 {ü}{{\"u}}1
    {Ä}{{\"A}}1 {Ë}{{\"E}}1 {Ï}{{\"I}}1 {Ö}{{\"O}}1 {Ü}{{\"U}}1
    {â}{{\^a}}1 {ê}{{\^e}}1 {î}{{\^i}}1 {ô}{{\^o}}1 {û}{{\^u}}1
    {Â}{{\^A}}1 {Ê}{{\^E}}1 {Î}{{\^I}}1 {Ô}{{\^O}}1 {Û}{{\^U}}1
    {œ}{{\oe}}1 {Œ}{{\OE}}1 {æ}{{\ae}}1 {Æ}{{\AE}}1 {ß}{{\ss}}1
    {ç}{{\c c}}1 {Ç}{{\c C}}1 {ø}{{\o}}1 {å}{{\r a}}1 {Å}{{\r A}}1
    {€}{{\EUR}}1 {£}{{\pounds}}1
}

\lstdefinelanguage{iPython}{
    morekeywords={access,and,break,class,continue,def,del,elif,else,except,exec,finally,for,from,global,if,import,in,is,lambda,not,or,pass,print,raise,return,try,while},%
    %
    morekeywords=[2]{abs,all,any,basestring,bin,bool,bytearray,callable,chr,classmethod,cmp,compile,complex,delattr,dict,dir,divmod,enumerate,eval,execfile,file,filter,float,format,frozenset,getattr,globals,hasattr,hash,help,hex,id,input,int,isinstance,issubclass,iter,len,list,locals,long,map,max,memoryview,min,next,object,oct,open,ord,pow,property,range,raw_input,reduce,reload,repr,reversed,round,set,setattr,slice,sorted,staticmethod,str,sum,super,tuple,type,unichr,unicode,vars,xrange,zip,apply,buffer,coerce,intern},%
    sensitive=true,%
    morecomment=[l]\#,%
    morestring=[b]',%
    morestring=[b]",%
    morestring=[s]{'''}{'''},
    morestring=[s]{"""}{"""},
    morestring=[s]{r'}{'},
    morestring=[s]{r"}{"},%
    morestring=[s]{r'''}{'''},%
    morestring=[s]{r"""}{"""},%
    morestring=[s]{u'}{'},
    morestring=[s]{u"}{"},%
    morestring=[s]{u'''}{'''},%
    morestring=[s]{u"""}{"""},%
    %
    literate=
    {á}{{\'a}}1 {é}{{\'e}}1 {í}{{\'i}}1 {ó}{{\'o}}1 {ú}{{\'u}}1
    {Á}{{\'A}}1 {É}{{\'E}}1 {Í}{{\'I}}1 {Ó}{{\'O}}1 {Ú}{{\'U}}1
    {à}{{\`a}}1 {è}{{\`e}}1 {ì}{{\`i}}1 {ò}{{\`o}}1 {ù}{{\`u}}1
    {À}{{\`A}}1 {È}{{\'E}}1 {Ì}{{\`I}}1 {Ò}{{\`O}}1 {Ù}{{\`U}}1
    {ä}{{\"a}}1 {ë}{{\"e}}1 {ï}{{\"i}}1 {ö}{{\"o}}1 {ü}{{\"u}}1
    {Ä}{{\"A}}1 {Ë}{{\"E}}1 {Ï}{{\"I}}1 {Ö}{{\"O}}1 {Ü}{{\"U}}1
    {â}{{\^a}}1 {ê}{{\^e}}1 {î}{{\^i}}1 {ô}{{\^o}}1 {û}{{\^u}}1
    {Â}{{\^A}}1 {Ê}{{\^E}}1 {Î}{{\^I}}1 {Ô}{{\^O}}1 {Û}{{\^U}}1
    {œ}{{\oe}}1 {Œ}{{\OE}}1 {æ}{{\ae}}1 {Æ}{{\AE}}1 {ß}{{\ss}}1
    {ç}{{\c c}}1 {Ç}{{\c C}}1 {ø}{{\o}}1 {å}{{\r a}}1 {Å}{{\r A}}1
    {€}{{\EUR}}1 {£}{{\pounds}}1
    {^}{{{\color{ipython_purple}\^{}}}}1
    {=}{{{\color{ipython_purple}=}}}1
    {+}{{{\color{ipython_purple}+}}}1
    {*}{{{\color{ipython_purple}$^\ast$}}}1
    {/}{{{\color{ipython_purple}/}}}1
    {+=}{{{+=}}}1
    {-=}{{{-=}}}1
    {*=}{{{$^\ast$=}}}1
    {/=}{{{/=}}}1,
    literate=
    *{-}{{{\color{ipython_purple}-}}}1
     {?}{{{\color{ipython_purple}?}}}1,
    identifierstyle=\color{black}\ttfamily,
    commentstyle=\color{ipython_cyan}\ttfamily,
    stringstyle=\color{ipython_red}\ttfamily,
    keepspaces=true,
    showspaces=false,
    showstringspaces=false,
    rulecolor=\color{ipython_frame},
    frame=single,
    frameround={t}{t}{t}{t},
    framexleftmargin=6mm,
    numbers=left,
    numberstyle=\tiny\color{halfgray},
    backgroundcolor=\color{ipython_bg},
    basicstyle=\scriptsize,
    keywordstyle=\color{ipython_green}\ttfamily,
}

\lstnewenvironment{python}[1][]
{
\pythonstyle
\lstset{#1}
}
{}

\begin{lstlisting}[language=iPython]
def search_cards(q=None):
    api_url = f"https://api.pokemontcg.io/v2/cards"
    querystring = {'q': q, }
    assert q is not None, 'Missing required parameter: q'
    
    response = requests.get(url=api_url, params=querystring, timeout=50, verify=False)
    if response.status_code != 200:
        response2 = requests.get(url=api_url, timeout=50) # in case API can't handle redundant params
        response = response2
    return response
    # print(response.json())

if __name__ == '__main__':
    r = search_cards(q='''name:gardevoir''')
    r_json = None
    try:
        r_json = r.json()
    except:
        pass
    import json
    result_dict = dict()
    result_dict['status_code'] = r.status_code
    result_dict['text'] = r.text
    result_dict['json'] = r_json
    result_dict['content'] = r.content.decode("utf-8")
\end{lstlisting}

\section{Tool Error Types and Causes} \label{app:error-validation}
Errors are categorized below:
\begin{itemize}
    \item \textbf{Incomplete URL:} Occasionally, the URL of a tool is incomplete, resulting in failed requests. These errors can be classified into two subtypes: \textbf{Missing Endpoint Path} or \textbf{Missing Base URL}. By examining the corresponding API documentations, we found that endpoint paths were usually provided, but base URLs were often missing.
    \item \textbf{Request Errors:} Request errors are complex and challenging to diagnose, as status codes alone do not clearly indicate whether the issue originates from the server or client side. A status code of 200 ("OK") guarantees valid communication between the client and server. To further validate the response content in such cases, we use an GPT-4o based evaluator (Appendix \ref{app:validation-prompt}). If the evaluator returns "pass", the tool is labeled as \textbf{Passed Validation}; otherwise, it is labeled as \textbf{Failed Validation}. For other status codes or unexpected cases, the tool is classified under \textbf{Abnormal Response}.
    \item \textbf{Incorrect Parameter Values:} If a tool throws an exception that does not fall into any of the previously mentioned scenarios, it indicates that the tool was very likely called with invalid parameters. Specifically:
    \begin{itemize}
        \item If a required parameter is missing, the error is classified as \textbf{No Parameter Value}.
        \item If all required parameters are provided but the tool still fails, the issue likely stems from an incorrect example value, and the error is classified as \textbf{Wrong Parameter Value}.
    \end{itemize}
\end{itemize}

For all error types other than \textbf{Passed Validation}, we group them into four main categories: \textbf{C1} Missing API Documentation Details, \textbf{C2} Incorrectly Extracted URL Path, \textbf{C3} Incorrect Parameter Values, and \textbf{C4} Server-Side Errors. For each category, we provide a range of possible error diagnosis, from the most conservative to the most aggressive (see Appendix \ref{app:error-validation}). 

\begin{table}[htbp]
\resizebox{\linewidth}{!}{
\begin{tabular}{lll}
\toprule
\multicolumn{1}{c}{\textbf{Category}}             & \multicolumn{1}{c}{\textbf{Conservative Estimate}}                                                 & \multicolumn{1}{c}{\textbf{Aggressive Estimate(Additional Terms Only)}}                          \\
\cmidrule(rl){1-1} \cmidrule(rl){2-2} \cmidrule(rl){3-3}
Missing API Documentation Details & 0                                                                                         & Missing Base URL+No Parameter Value   \\
Incorrectly Extracted URL Path    & Missing Endpoint Path                                           & Missing Base URL                                              \\
Incorrect Parameter Values   & Wrong Parameter Value+Failed Validation & No Parameter Value+Abnormal Response  \\
Server-Side Error                              & 0                                                                                         & Failed Validation+Abnormal Response \\
\bottomrule
\end{tabular}
}
\caption{Estimation of error causes}
\label{app:tab:error-estimation}
\end{table}

\newpage
\section{APILLAMA training} \label{app:apillama-training}
\begin{figure}[htbp]
    \centering
    \includegraphics[width=0.5\linewidth]{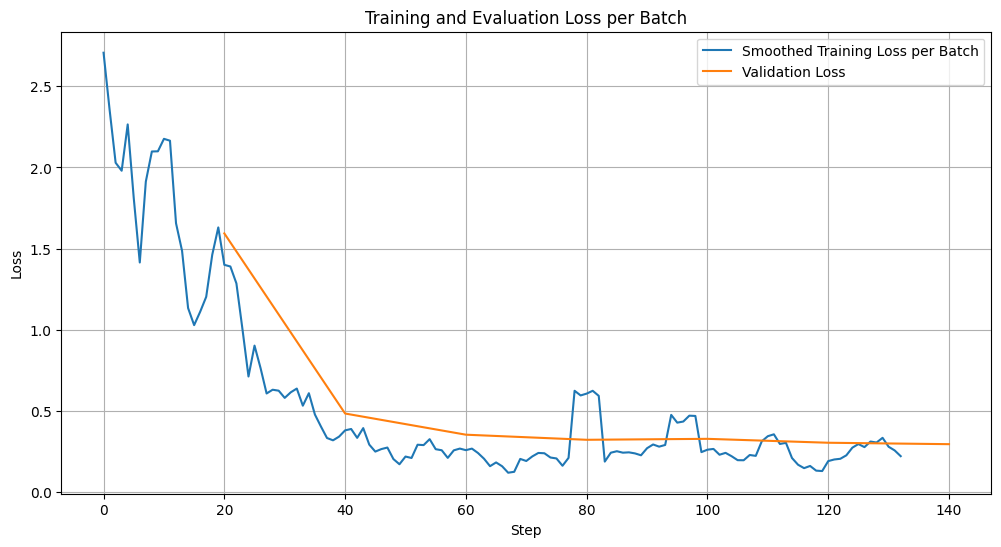}
    \caption{Training loss of APILLAMA. We applied moving average of window size 5 to smooth the training curve. Validation result is produced in every 20 steps.}
    \label{app:fig:training-loss}
\end{figure}

APILLAMA is fine-tuned on the basis of \verb|meta-llama/Meta-Llama-3-8B-Instruct|. The default configuration has hidden dimension $4,096$ and context length $8,192$. We extend the context length to $10,240$ to support longer API documentation input. A trainable soft prompt with $20$ virtual tokens is added, with the weight initialzed by the instruction prompt "You will be given an API documentation. Extract the API endpoints and output in JSON format.". To wrap up, APILLAMA has a total of $20\times4096=81920$ trainable parameters. We use an AdamW optimizer with learning rate $0.001$, a linear scheduler with warm-up steps $10$ and negative log likelihood loss for training. We train the model with one epoch, $149$ data points. Figure \ref{app:fig:training-loss} shows the training loss curve for APILLAMA.

\section{Tool Validation Errors}
Table \ref{app:tab:tool-errors} displays the errors occurred in Section \ref{sec:experiment-2}.

\begin{table}[htbp]
\resizebox{\linewidth}{!}{
\begin{tabular}{lccccll}
\toprule
                        & \multicolumn{6}{c}{Error Type}                                                                                                                                                                                                                                                                               \\ \cmidrule(rl){2-7}
                       & \multicolumn{1}{c}{Missing Endpoint Path}   & \multicolumn{1}{c}{Missing Base URL}                 & \multicolumn{1}{c}{Failed Validation}                  & \multicolumn{1}{c}{Abnormal Response}  & No Parameter Value              & Wrong Parameter Value            \\ \midrule
Ground Truth                                           & 0                                   & 4                                         & 9                                         & 23                           & \multicolumn{1}{c}{14 } & \multicolumn{1}{c}{10 }  \\
LLaMA3+one shot                                          & 0                                  & 4                                            & 0                                              & 2                                & \multicolumn{1}{c}{0 }   & \multicolumn{1}{c}{4}  \\
GPT3.5+one shot                                         & 0                                  & 1                                            & 4                                               & 5                           & \multicolumn{1}{c}{0}   & \multicolumn{1}{c}{1}  \\
GPT3.5 structured mode                                  & 0                                    & 9                                          & 4                                               & 21                             & \multicolumn{1}{c}{14} & \multicolumn{1}{c}{10} \\
\textbf{APILLAMA(Ours)}                                  & 0                           & 0                                           & 4                                               & 20                            & \multicolumn{1}{c}{0}   & \multicolumn{1}{c}{16} \\ \bottomrule 
\end{tabular}
}
\resizebox{\linewidth}{!}{
\begin{tabular}{lccccc}
\toprule
\textbf{}                           & \multicolumn{4}{c}{Error Cause}        \\ \cmidrule(rl){2-5}
                       & \multicolumn{1}{c}{Missing API DOC Details} & \multicolumn{1}{c}{Incorrectly Extracted URL Path} & \multicolumn{1}{c}{Incorrect Parameter Values} & \multicolumn{1}{c}{Server-side Errors} \\ \midrule
Ground Truth                                            & 0-18                            & 0-4                                        & 19-56                                      & 0-32                   \\
LLaMA3+one shot                                          & 0-4                              & 0-4                                       & 4-6                                        & 0-2                        \\
GPT3.5+one shot                                        & 0-1                               & 0-1                                        & 5-10                                     & 0-9                       \\
GPT3.5 structured mode                                  & 0-23                               & 0-9                                       & 14-47                                      & 0-25                       \\
\textbf{APILLAMA(Ours)}                                & 0-0                               & 0-0                                        & 20-36                                     & 0-24                  \\ \bottomrule
\end{tabular}
}
\caption{Analysis of errors in generated tools. The top table shows the count of tools where certain each error type occurs. The bottom table shows an estimation of the cause for the errors, from the most conservative estimate to the most aggressive estimate.}
\label{app:tab:tool-errors}
\end{table}

\section{Parameters in Glycoscience APIs}
\begin{table}[H]
\centering

\renewcommand{\arraystretch}{1.5}
\begin{tabular}{ccc}
\toprule
\multicolumn{3}{c}{\textbf{String Representation}}  \\ \cmidrule(rl){1-3}
\multicolumn{1}{c}{\textbf{IUPAC Condensed}} & \multicolumn{2}{c}{Fuc(a1-2)Gal(b1-3){[}Fuc(a1-4){]}GlcNAc(b1-}     \\
\multicolumn{1}{c}{\textbf{GLYCAM}}          & \multicolumn{2}{c}{LFucpa1-2DGalpb1-3{[}LFucpa1-4{]}DGlcpNAcb1-OH}   \\ \bottomrule \toprule
\multicolumn{2}{c}{\textbf{Database ID}} & \multicolumn{1}{c}{\textbf{Glycan Name}} \\ \cmidrule(rl){1-2} \cmidrule(rl){3-3}
\multicolumn{1}{c}{\textbf{GlyToucan ID}} & \multicolumn{1}{c}{G00048MO} & \multirow{2}{*}{Lewis b} \\
\multicolumn{1}{c}{\textbf{PubChem ID}} & \multicolumn{1}{c}{45480569} & \multicolumn{1}{c}{} \\
\bottomrule
\end{tabular}
\caption{Example of different representations of Glycan \textit{Lewis b}. The table shows various ways to represent the glycan, demonstrating the complexity of glycan reference.}
\label{app:tab:glycanid}
\end{table}

\begin{table}[H]
\centering

\begin{tabular}{ccc}
\toprule
\textbf{Database} & \textbf{Reference ID} & \textbf{Primary Usage}                                                            \\ 
\midrule
GlyTouCan         & GlyTouCan ID          & \begin{tabular}[c]{@{}c@{}}Glycan Structure\\ GlyTouCan ID for API Calling\end{tabular} \\ \midrule
KEGG GLYCAN       & KEGG ID               & \begin{tabular}[c]{@{}c@{}}KEGG Pathyway\\ Reaction\end{tabular}                        \\ \midrule
GlyGen            & GlyTouCan ID          & \begin{tabular}[c]{@{}c@{}}Publication\\ Cross Reference\end{tabular}                   \\ \midrule
O-GlcNAc          & UniProtKB ID          & Protein O-GlcNAcylation Data                                                            \\ \midrule
PubChem           & PubChem CID           & Chemical Properties                                                                     \\ \midrule
Unilectin         & Unilectin ID          & Get Lectin and Ligand Information                                                       \\ 
\bottomrule
\end{tabular}

\caption{Glycan databases covered in glyco agent}
\label{app:tab:glycodataset}
\end{table}

\section{LLM Prompts}
\subsection{API Response Validation Prompt} \label{app:validation-prompt}
    
\begin{lstlisting}[language=iPython]
'''
Decide if the following API response is an information or an error message.

API Description:
{description}

API Response:
{response}
'''
\end{lstlisting}

\subsection{Parameter Generation Prompt}

\begin{lstlisting}[language=iPython]
'''
You will be provided with the information of an API and its parameters. The example values of the parameters are missing. You need to guess the parameter values.
You may have failed severl times before. If you guess with similar values, you may fail again. Please be innovative and try different values and formats.

Your previous failed guesses:
***history start
{history}
***history end

API Description:
{description}

Parameter Description:
{param_description}

Your Guess:
'''

# Output Class Structure
class Parameter(BaseModel):
    parameter_key: str
    parameter_guess: str = Field(..., description="The guessed values of the parameter.")

class ParameterList(BaseModel):
    parameters: List[Parameter] = Field(..., min_items=1, description="The list of parameters and their guesses.")
\end{lstlisting}

\subsection{API Documentation Classification Prompt} \label{app:doc-classification-prompt}
\begin{lstlisting}[language=iPython]
'''
You need to group the API documentation with the following standards:

Fully Organized: The documentation follows a well defined template, most likely to be from an API platform. It is well-structured, clear, and easy to understand. It includes detailed descriptions, example code, and explanations of how to use the API.
Semi-Organized: Lacks some structure, but still includes most of the necessary information. It may be missing some examples or descriptions, making it slightly more difficult to understand how to use the API.
Unorganized: Missing example or description, or the structure is unclear, making it difficult to understand how to use the API.

===
API Documentation:
{API_DOC}
'''

# Output Class Structure
class Classification(BaseModel):
    analysis: str = Field(..., description="The analysis of the API documentation. Make it within 300 characters.")
    category: str = Field(..., enum=["Fully Organized", "Semi-Organized", "Unorganized"])
\end{lstlisting}

\AddToHook{enddocument/afteraux}{%
\immediate\write18{
cp output.aux ijcai25.aux
}%
}
\end{document}